\documentclass{article} 
\usepackage{icomp2024_conference,times}

\usepackage{amsmath,amsfonts,bm}

\def\eqref#1{equation~\ref{#1}}

\def\1{\bm{1}}

\def\vx{{\bm{x}}}

\def\vz{{\bm{z}}}

\def\evx{{x}}

\def\evz{{z}}

\DeclareMathAlphabet{\mathsfit}{\encodingdefault}{\sfdefault}{m}{sl}
\SetMathAlphabet{\mathsfit}{bold}{\encodingdefault}{\sfdefault}{bx}{n}

\usepackage{hyperref}
\usepackage{url}
\usepackage{url}
\usepackage{booktabs}
\usepackage{amsfonts}
\usepackage{nicefrac}
\usepackage{longtable}
\usepackage{microtype}
\usepackage{colortbl}

\usepackage{amsmath}
\usepackage{amssymb}
\usepackage{caption}
\usepackage{enumitem}
\usepackage{float}
\usepackage{multirow}
\usepackage{tcolorbox}
\usepackage{xcolor}
\usepackage{tikz}
\usepackage{ulem}
\usetikzlibrary{positioning, arrows.meta, decorations.pathreplacing}
\usetikzlibrary{shapes.geometric, arrows, positioning}

\usetikzlibrary{arrows.meta}
\title{Strategic inputs: feature selection from game-theoretic perspective}

\author{Chi~Zhao, \& Jing Liu \& Elena Parilina\\
Saint Petersburg State University,\\
7/9 Universitetskaya nab.,\\
Saint Petersburg, 199034, Russia\\
\texttt{chi.zhao@spbu.ru},
\texttt{st082130@student.spbu.ru},
\texttt{e.parilina@spbu.ru}
}

\icompfinalcopy 
\begin{document}

\maketitle

\begin{abstract}
	The exponential growth of data volumes has led to escalating computational costs in machine learning model training. However, many features fail to contribute positively to model performance while consuming substantial computational resources. This paper presents an end-to-end feature selection framework for tabular data based on game theory. We formulate feature selection procedure based on a cooperative game where features are modeled as players, and their importance is determined through the evaluation of synergistic interactions and marginal contributions. The proposed framework comprises four core components: sample selection, game-theoretic feature importance evaluation, redundant feature elimination, and optimized model training. Experimental results demonstrate that the proposed method achieves substantial computation reduction while preserving predictive performance, thereby offering an efficient solution of the computational challenges of large-scale machine learning. The source code is available at \url{https://github.com/vectorsss/strategy_inputs}.
\end{abstract}

\section{Introduction}

A growth of data has created significant challenges for machine learning
practitioners: it significantly increases computational cost and model
performance. As datasets grow in both size and dimensionality, there is an
actual need of creation of efficient feature selection methods that can reduce
dimensionality while maintaining predictive performance. Feature selection is a
crucial step in the machine learning pipeline since it directly impacts both
model performance and training efficiency.

Development of explainable AI (XAI) has introduced new paradigms for feature
selection based on feature importance evaluation. These XAI-based methods can
be divided into two main groups: game-theoretic and non game-theoretic.

Non game-theoretic methods, such as Feature Importance (FI)
\citep{breiman2001random}, Permutation Feature Importance (PFI)
\citep{altmann2010permutation}, and Local Interpretable Model-Agnostic
Explanations (LIME) \citep{ribeiro2016should} are widely used in practice due
to their computational efficiency and ease of implementation. However, these
methods typically evaluate feature importance independently of other features,
i.e., they cannot capture complex interactions between features.

Game-theoretic methods, particularly, those based on the Shapley value, provide
a more comprehensive framework for feature importance evaluation. The SHAP
method (SHapley Additive exPlanations) \citep{lundberg2017unified} provides a
unified approach with various implementations including KernelSHAP,
SamplingSHAP, etc. Recently developed ShapG method
\citep{zhao2024shapgnewfeatureimportance,zhao2025complete}, inspired by Shapley
value-based centrality measures on graphs
\citep{zhao2024centralitymeasuresopiniondynamics}, demonstrates improved
computational efficiency compared to traditional Shapley-based methods. A
sparsity-based deterministic method for the Shapley value approximation
presented in \citet{erofeeva2025novel} introduces a novel compressed-sensing
perspective on this problem. However, as the number of players (features)
increases, the associated time and memory requirements grow rapidly, which may
limit its applicability in high-dimensional settings. These game-theoretic
approaches evaluate feature importance based on cooperative interactions among
features, enabling more detailed understanding of feature contributions. A
detailed description and benchmark results for each method can be found in
\citep{zhao2025shapley}.

Despite their theoretical advantages, game-theoretic methods face significant
computational challenges. The computational cost increases exponentially with
the number of features when using the Shapley value to measure their
importance. Both game-theoretic and non game-theoretic methods exhibit a linear
relationship between data scale (number of observations) and computational cost
for feature importance evaluation. Therefore, selecting a representative subset
of observations can significantly reduce computational costs for feature
importance evaluation. In the paper \citet{axiotis2024data}, the authors
proposed a clustering-based sensitivity sampling method to select diverse
subsets of observations for efficient learning. In this work, we adopt a
simpler clustering-based method that emphasizes balanced coverage of the data
distribution, ensuring the sampled subset reflects both global and local
diversity without over-emphasizing highly influential observations. The latter
may lead to inaccurate feature importance estimators.

The main contributions of this paper are as follows:

\begin{enumerate}
	\item
	      We propose a novel clustering-based sampling method that ensures balanced (representative) coverage of the data distribution.
	\item
	      We move the feature importance evaluation from the training stage to the
	      prediction stage, allowing efficient feature selection without retraining
	      the model.
	\item
	      We propose a feature selection pipeline that integrates our sampling method and importance evaluation into a unified framework and benchmark its
	      performance against state-of-the-art methods.
	\item
	      We demonstrate the effectiveness of the proposed method on various tabular
	      datasets, showing that it accelerates the model training process without
	      significant loss in model performance.
\end{enumerate}

The rest of the paper is organized as follows: in Section
\ref{sec:problem_formulation} we describe the problem and the way how we are
solving it. It includes data sampling procedure (Section \ref{sec:diversity}),
feature selection (Sections \ref{sec:feature_selection} and
\ref{sec:pipeline}). Experiments on ten datasets are described in Section
\ref{sec:experiments}. We briefly conclude and present our future work in
Section \ref{sec:conclusion}.

\section{Problem formulation}
\label{sec:problem_formulation}

In this section, we describe the problem of feature selection from a
game-theoretic perspective and the proposed pipeline for this task.

We consider a supervised learning setting with dataset $D = \{(\vx^{(i)},
	y^{(i)})\}_{i=1}^n$ consisting of $n$ observations, where $\vx^{(i)} =
	(\evx_1^{(i)}, \ldots, \evx_M^{(i)}) \in \mathbb{R}^M$ represents the $i$th
observation of a feature vector consisting of $M$ features, and $y^{(i)}\in
	\mathbb{R}$ is the $i$th observation of a target variable. We assume that the
dataset is ``cleaned'' and preprocessed, meaning that there are no missing
values, no outliers, etc. Otherwise, it is necessary to perform data
preprocessing before feature selection.

There is a linear relationship between the data scale (number of observations)
and the computational cost of feature importance evaluation. Therefore, we
develop a two-stage feature selection pipeline, in which the first stage
involves selecting a subset of representative observations from the original
dataset (Section \ref{sec:diversity}), and the second stage evaluates the
feature importance based on the selected sample (Section
\ref{sec:feature_selection}).

\subsection{Data sampling}
\label{sec:diversity}

We have dataset $D$ described above with $n$ observations and we aim in
selecting a representative of sample size $s$, i.e., subset $S\subseteq D$ with
$|S| = s<n$. The term ``representative'' means that the selected subset $S$ is
able to express the original dataset $D$ in such a way that every observation
from $D$ has a representative in $S$ within a bounded distance.

For this purpose, we propose a diversity-based sampling method, which includes
the following steps:

\begin{enumerate}
	\item[\textbf{Step 1}]
	      For a given $k$, find cluster centers $C = \{c_1, \ldots, c_k\}$ (we use $k=\lfloor \sqrt{s} \rfloor$ as default under the assumption that data is balanced, i.e. in each cluster $C_i$ the number of observations is, roughly speaking, equal) minimizing the within-cluster variance:
	      \begin{equation*}
		      C^* = \arg\min_{C} \sum_{x \in D} \min_{c \in C} \|x - c\|^2.
	      \end{equation*}
	      This can be solved by the $k$-means++ algorithm (see \citet{lloyd1982least, arthur2006k}). By solving this minimization problem, we obtain the set of clusters $\{C_1,\ldots,C_k\}$.

	\item[\textbf{Step 2}]
	      For each cluster $C_i$, $i=1,\ldots,k$, select $s_i$ observations, where
	      \begin{equation*}
		      s_i =
		      \begin{cases}
			      \lfloor \frac{s}{k} \rfloor + 1, & \text{if } i \leq (s \mod k), \\
			      \lfloor \frac{s}{k} \rfloor,     & \text{otherwise}.
		      \end{cases}
	      \end{equation*}

	      Since we aim to select $s$ observations in total, we need to distribute them
	      among $k$ clusters. Therefore, we need to select $\lfloor s/k \rfloor$
	      observations from each cluster $C_i$, $i=1,\ldots,k$, and then distribute the
	      remaining $(s \mod k)$ observations among the first $(s \mod k)$ clusters. This
	      ensures that the total number of observations selected is equal to $s$.

	      For cluster $C_i$ with observations sorted by increasing distance to center
	      $c_i$, let $\pi_i: \{1, \ldots, |C_i|\} \rightarrow C_i$ be a rank of the
	      $i$-th point in cluster $C_i$, such that:
	      \begin{equation*}
		      \|\pi_i(1) - c_i\| \leq \|\pi_i(2) - c_i\| \leq ... \leq \|\pi_i(|C_i|) - c_i\|.
	      \end{equation*}

	      Select $s_i$ observations from $C_i$ with uniform intervals:
	      \begin{equation*}
		      S_i = \{\pi_i(j \cdot \Delta_i) : j \in \{0, 1, ..., s_i-1\} \text{ and } j \cdot \Delta_i < |C_i|\},
	      \end{equation*}
	      where $\Delta_i = \max(1, \lfloor|C_i|/s_i\rfloor)$ is the distance between selected observations.
\end{enumerate}

This approach ensures both global diversity (through clustering) and local
diversity (through evenly-spaced sampling within each cluster), providing a
computationally efficient method that approximates maximizing the sum of
pairwise distances in the selected subset.

\subsection{Feature selection}
\label{sec:feature_selection}

Assume that we have a model $f: \mathbb{R}^M \rightarrow \mathbb{R}$, which
receives an input vector $\vx = (\evx_1, \ldots, \evx_M) \in \mathbb{R}^M$,
representing an observation of $M$ features, and gives the output $f(\vx)$.

The feature selection problem can be formulated from a cooperative game
perspective with the goal in identifying a subset of features that maximizes
the predictive performance of a model. In other words, we can identify a subset
of features that have low impact on the model performance and this subset
should be removed from consideration. We formulate this problem using
cooperative game theory, where each feature is treated as a player whose
contributions to model performance can be measured.

A cooperative game is described by the pair $(\mathcal{M}, v)$, where
$\mathcal{M}=\{\mathcal{F}_1,\ldots,\mathcal{F}_M\}$ represents the set of
players/features, and $v: 2^{\mathcal{M}} \rightarrow \mathbb{R}$ is a
characteristic function assigning a ``strength'' to any coalition of players
(i.e., any subset $\mathcal{S} \subset \mathcal{M}$, with $S=|\mathcal{S}|$).
Specifically, the value $v(\mathcal{S})$ quantifies the ``strength'',
predictive power, or contribution of coalition $\mathcal{S}$.

The first step in feature selection procedure is to evaluate importance of all
features, and the second step is to remove features with low importance. To
evaluate $v(\mathcal{S})$ for any coalition, we define the hybrid input
$\vz^{(\mathcal{S})}=(\evz_{1}^{(\mathcal{S})}, \ldots,
	\evz_{M}^{(\mathcal{S})})\in \mathbb{R}^M$ by
\begin{equation}
	\evz_{i}^{(\mathcal{S})} =
	\begin{cases}
		\evx_i, & \text{if } i \in \mathcal{S},    \\
		r_i,    & \text{if } i \notin \mathcal{S}.
	\end{cases}
\end{equation}
If a feature belongs to coalition
$\mathcal{S}$, its values from dataset $D$ are collected. Otherwise, the
reference value of a feature is calculated based on dataset $D$. Given dataset $D$, we can obtain a reference value
$r_i$ for each feature $\mathcal{F}_i$, calculated as a statistic from the
feature observations (such as mean, median, or mode).
This hybrid input $\evz_{i}^{(\mathcal{S})}$ replaces features absent in coalition $\mathcal{S}$ with
their reference values while preserving the values of features in
$\mathcal{S}$.

There are several methods to define characteristic function $v(\mathcal{S})$:
\begin{enumerate}
	\item
	      \textbf{Method 1. Model output}:
	      \begin{equation}
		      \label{eq:model_output}
		      v(\mathcal{S}) = f(\vz^{(\mathcal{S})}),
	      \end{equation}
	      which is the most straightforward approach used in many XAI methods based on cooperative game theory. The change in model output when a feature is present versus it is absent indicates a feature importance. If the output does not change, the feature is unimportant. The magnitude of change reflects the degree of importance.

	\item
	      \textbf{Method 2. Sample-level performance-based metric}: characteristic function $v(\mathcal{S})$ is defined as follows:
	      \begin{equation}
		      \label{eq:sample_level_performance_metric}
		      v(\mathcal{S}) =
		      \begin{cases}
			      -\|f(\vz^{(\mathcal{S})}) - y\|^2,                      & \text{(regression task)}     \\
			      2\cdot \mathbf{1}_{\{f(\vz^{(\mathcal{S})}) = y\}} - 1, & \text{(classification task)}
		      \end{cases}
	      \end{equation}

	      where $\mathbf{1}_{\{\cdot\}}$ is the indicator function. For classification,
	      this returns $+1$ for correct predictions and $-1$ for incorrect ones. Negative
	      values indicate features that harm model performance.

	\item
	      \textbf{Method 3. Global-model performance}: Similar to
	      \eqref{eq:sample_level_performance_metric}, but using dataset-wide metrics: $R^2$, negative MAE, or negative RMSE for regression tasks; accuracy or F1 score for classification tasks (\citet{zhao2024shapgnewfeatureimportance}).

\end{enumerate}

The first two methods provide sample-level metrics, while the third provides
global metrics. For sample-level metrics, we can obtain a global feature
importance by averaging across all samples. When using model output (Method 1),
we average the absolute values since both increases and decreases indicate
importance. Global performance metrics (Method 3) directly yields feature
importance scores without averaging.

Once the values of characteristic function $v(\mathcal{S})$ are defined using
one of the proposed methods, a challenge in cooperative game theory is in
determining a ``fair'' allocation of the total payoff $v(\mathcal{M})$ (the
value of the grand coalition) distributing its payoff among participating
players.

The Shapley value \citep{shapley1951notes} is a widely used solution of this
problem:
\begin{equation*}
	\phi_i = \sum_{\mathcal{S} \subseteq \mathcal{M} \setminus \{i\}} \frac{|\mathcal{S}|!(M-|\mathcal{S}|-1)!}{M!} \left(v(\mathcal{S} \cup \{i\}) - v(\mathcal{S})\right), i \in \mathcal{M}.
\end{equation*}

The Centre of the Imputation Set (CIS) Value \citep{driessen1991coincidence}
offers a more computationally efficient alternative:
\begin{equation*}
	CIS_i(v) = v(\{i\}) + M^{-1} \left[ v(\mathcal{M}) - \sum_{j \in \mathcal{M}} v(\{j\})\right], i \in \mathcal{M},
\end{equation*}
and it requires computation of only $M+1$
values of $v$ compared to $2^M$ values for the exact Shapley value.

To further improve computational efficiency, we adapt the graph-based sampling
approach from the ShapG method \citep{zhao2024shapgnewfeatureimportance}. ShapG
constructs a collaborator set $\mathcal{C}_i$ for each feature $i$, restricting
feature interactions to within these sets during evaluation. When this set is
not large, i.e. $|\mathcal{C}_i| < \ell$ (default $\ell=15$), the exact Shapley
value can be computed; otherwise, Monte Carlo approximation is used. We
integrate this approach into our pipeline without requiring model retraining,
significantly reducing computational costs.

In this paper, we focus primarily on the sample-level performance-based metric
for feature importance evaluation since it provides fine-grained insights into
feature contributions. Global model performance metrics are used as a
post-processing step to validate the overall effectiveness of the selected
features.

\subsection{Proposed Pipeline for Feature Selection}\label{sec:pipeline}

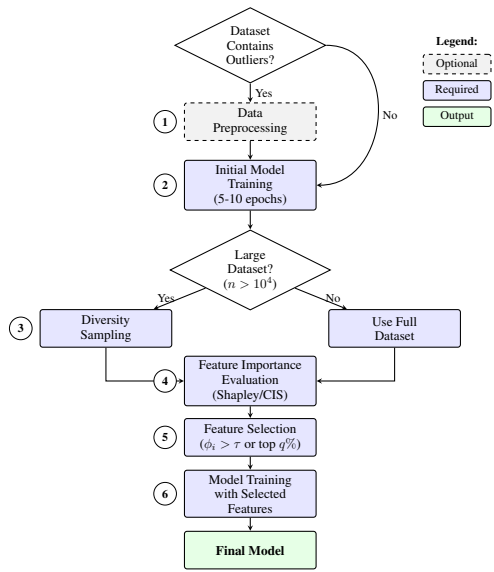
\begin{figure}[h]
	\centering
	\scalebox{0.5}{\begin{tikzpicture}[
		node distance=1.5cm,
		box/.style={rectangle, draw, thick, minimum width=3.5cm, minimum height=1cm, align=center, rounded corners=2pt},
		decision/.style={diamond, draw, thick, minimum width=2.5cm, minimum height=1cm, align=center, aspect=2},
		arrow/.style={->, thick, >=stealth},
		optional/.style={box, dashed, fill=gray!10},
		main/.style={box, fill=blue!10},
		label/.style={font=\footnotesize, text width=2.5cm, align=center}
	]

	\node[decision] (start) {Dataset\\Contains\\Outliers?};
	\node[optional, below=0.5cm of start] (preprocess) {Data\\Preprocessing};

	\node[main, below=0.5cm of preprocess] (train) {Initial Model\\Training\\(5-10 epochs)};

	\node[decision, below=0.5cm of train] (size) {Large\\Dataset?\\($n > 10^4$)};
	\node[main, below left=0.5cm and 1cm of size] (sample) {Diversity\\Sampling};
	\node[main, below right=0.5cm and 1cm of size] (direct) {Use Full\\Dataset};

	\node[main, below=1.2cm of size] (importance) {Feature Importance\\Evaluation\\(Shapley/CIS)};
	\node[main, below of=importance] (select) {Feature Selection\\($\phi_i > \tau$ or top $q\%$)};
	\node[main, below of=select] (retrain) {Model Training\\with Selected\\Features};
	\node[box, fill=green!10, below of=retrain] (final) {\textbf{Final Model}};

	\draw[arrow] (start) -- node[right, font=\footnotesize] {Yes} (preprocess);
	\draw[arrow] (start.east) .. controls +(2,0) and +(2,0) .. node[right, font=\footnotesize] {No} (train.east);
	\draw[arrow] (preprocess) -- (train);
	\draw[arrow] (train) -- (size);
	\draw[arrow] (size) -- node[left, font=\footnotesize] {Yes} (sample);
	\draw[arrow] (size) -- node[right, font=\footnotesize] {No} (direct);
	\draw[arrow] (sample) |- (importance);
	\draw[arrow] (direct) |- (importance);
	\draw[arrow] (importance) -- (select);
	\draw[arrow] (select) -- (retrain);
	\draw[arrow] (retrain) -- (final);

	\node[circle, draw, fill=white, font=\small\bfseries] at ([xshift=-0.5cm]preprocess.west) {1};
	\node[circle, draw, fill=white, font=\small\bfseries] at ([xshift=-0.5cm]train.west) {2};
	\node[circle, draw, fill=white, font=\small\bfseries] at ([xshift=-0.5cm]sample.west) {3};
	\node[circle, draw, fill=white, font=\small\bfseries] at ([xshift=-0.5cm]importance.west) {4};
	\node[circle, draw, fill=white, font=\small\bfseries] at ([xshift=-0.5cm]select.west) {5};
	\node[circle, draw, fill=white, font=\small\bfseries] at ([xshift=-0.5cm]retrain.west) {6};

	\node[optional, minimum width=1.8cm, minimum height=0.4cm] at (5.5, -0.5) (leg1) {\footnotesize Optional};
	\node[main, minimum width=1.8cm, minimum height=0.4cm] at (5.5, -1.2) (leg2) {\footnotesize Required};
	\node[box, fill=green!10, minimum width=1.8cm, minimum height=0.4cm] at (5.5, -1.9) (leg3) {\footnotesize Output};
	\node[above=0.05cm of leg1, font=\footnotesize\bfseries] {Legend:};
\end{tikzpicture}}
	\caption{Proposed pipeline for feature selection.}
	\label{fig:pipeline}
\end{figure}

Our feature selection pipeline consists of the following steps as shown in
Figure~\ref{fig:pipeline}.

\begin{enumerate}
	\item
	      \textbf{Data Preprocessing} [Optional]: If dataset contains outliers, apply
	      appropriate outlier detection and removal techniques. Otherwise, proceed with the next step.

	\item
	      \textbf{Initial Model Training}: Train an initial model on a representative subset for a limited number of epochs (e.g., 5-10 epochs) to obtain a preliminary model.

	\item
	      \textbf{Diversity Sampling}: For large datasets (e.g., $n > 10^4$ observations), apply the diversity
	      sampling method described in Section~\ref{sec:diversity} to select a representative
	      subset of observations for feature importance evaluation.

	\item
	      \textbf{Feature Importance Evaluation}: Using the trained initial model, compute feature
	      importance scores for all features using the characteristic functions defined in
	      Section~\ref{sec:feature_selection} (e.g., model output or negative squared error).

	\item
	      \textbf{Feature Selection}: Remove features with importance scores below a threshold
	      $\tau$, or only keep the top $q\%$ of features. Specifically, retain feature $i$ if
	      $\phi_i > \tau$ or $\phi_i$ is in the top $q\%$ of overall importance scores, where
	      $\phi_i$ is the Shapley value or CIS value $i$th component.

	\item
	      \textbf{Model training}: Train a final model using only the selected features on the
	      complete dataset (or the sampled subset for very large datasets).
\end{enumerate}

\section{Experiments}\label{sec:experiments}

In this section, we evaluate how our proposed feature selection pipeline
affects the performance of deep-learning models on tabular data, focusing on
MLP and TabM \citep{gorishniy2024tabm}, the current state-of-the-art on TabRed
benchmarks \citep{rubachev2024tabred}.

\subsection{Experimental setup}

\subsubsection{Models}
We evaluate four model configurations based on the TabM benchmark suite:

\begin{enumerate}
	\item
	      \textbf{MLP}: A multilayer perceptron with fully connected layers\footnote{We use the same architecture as in \citet{gorishniy2024tabm}, the number of layers varies according to the dataset.}, ReLU activations, and task-specific output heads.

	\item
	      \textbf{TabM}: A parameter-efficient ensemble model for tabular data that trains multiple MLPs simultaneously while sharing most parameters between them, achieving state-of-the-art performance \citep{gorishniy2024tabm}.

	\item
	      \textbf{MLP+PLE}: MLP enhanced with PLE (Piecewise Linear Embedding, see \cite{gorishniy2022embeddings}) for numerical features, which splits continuous values into segments and learns separate representations for each segment, enabling better modeling of non-linear relationships.

	\item
	      \textbf{TabM+PLE}: TabM with PLE, combining attention mechanisms with sophisticated numerical encoding.
\end{enumerate}

\subsubsection{Configuration}

\begin{table}[h]
	\centering
	\caption{Experiment Configuration}
	\scalebox{0.85}{\begin{tabular}{ll}
	\hline
	\textbf{Parameter}                                                 & \textbf{Value}                                                          \\
	\hline
	Feature importance method                                          & ShapG and PFI                                                           \\
	Characteristic function                                            & Sample-level performance (Eq.~\ref{eq:sample_level_performance_metric}) \\
	Feature retention (top-$q\%$)                                      & 80\%                                                                    \\
	Sampling threshold $\ell$                                          & 5                                                                       \\
	Number of seeds                                                    & 3                                                                       \\
	Training strategy                                                  & Early stopping (patience=16)                                            \\
	Distance metric for collaboration set $\mathcal{C}_i$ construction & 1 - cosine similarity                                                   \\
	Evaluation metrics                                                 & Accuracy (classification), $R^2$ (regression)                           \\
	\hline
\end{tabular}}
\end{table}

We compare baseline models (all the features included) against models with
feature selection (FS), reporting mean $\pm$ std of model performance and
training time across 3 seeds.

\subsection{Datasets}

In total, we use 10 datasets from \citep{gorishniy2024tabm}.

\begin{table}[h]
	\caption{Dataset configuration from \citet{gorishniy2024tabm}}
	\setlength\tabcolsep{2.2pt}
	\centering
	\scalebox{0.8}{\begin{tabular}{lccccccccccl}
	\hline
	\textbf{Name}       & \textbf{Abbr} & \textbf{\#Train} & \textbf{\#Validation} & \textbf{\#Test} & \textbf{\#Num} & \textbf{\#Bin} & \textbf{\#Cat} & \textbf{Task type} & \textbf{Batch size} \\
	\hline
	Adult               & AD            & 26\,048          & 6\,513                & 16\,281         & 6              & 1              & 8              & Binclass           & 256                 \\
	Black Friday        & BF            & 106\,764         & 26\,692               & 33\,365         & 4              & 1              & 4              & Regression         & 512                 \\
	California Housing  & CH            & 13\,209          & 3\,303                & 4\,128          & 8              & 0              & 0              & Regression         & 256                 \\
	Churn Modelling     & CM            & 6\,400           & 1\,600                & 2\,000          & 7              & 3              & 1              & Binclass           & 128                 \\
	Covertype           & CO            & 371\,847         & 92\,962               & 116\,203        & 10             & 4              & 1              & Multiclass         & 1024                \\
	Diamond             & DI            & 34\,521          & 8\,631                & 10\,788         & 6              & 0              & 3              & Regression         & 512                 \\
	Higgs Small         & HS            & 62\,751          & 15\,688               & 19\,610         & 28             & 0              & 0              & Binclass           & 512                 \\
	House 16H           & HO            & 14\,581          & 3\,646                & 4\,557          & 16             & 0              & 0              & Regression         & 256                 \\
	Microsoft           & MI            & 723\,412         & 235\,259              & 241\,521        & 131            & 5              & 0              & Regression         & 1024                \\
	Otto Group Products & OT            & 39\,601          & 9\,901                & 12\,376         & 93             & 0              & 0              & Multiclass         & 512                 \\
	\hline
\end{tabular}}
	\label{tab:dataset_stats}
\end{table}

\subsection{Implementation details}

\textbf{Training.} We use the TabM benchmark suite with hyperparameters from
\citet{gorishniy2024tabm}, tuned via Optuna \citep{akiba2019optuna}.

\noindent\textbf{Feature importance.} Since many datasets in our benchmark
have more than 20 features, most existing feature-importance XAI methods are nearly
infeasible in this setting. Thus, we focus on two methods for feature importance
evaluation:
\begin{enumerate}
	\item
	      \textbf{ShapG method}:
	      We adapt ShapG \citep{zhao2024shapgnewfeatureimportance, zhao2025complete} by
	      moving evaluation from training to prediction stage, avoiding model retraining
	      for each feature subset. Following \citet{zhao2025complete}'s improved
	      strategy, we construct collaboration sets $\mathcal{C}_i$ through a graph-based
	      approach: (1) rank features by their correlation with the target using cosine
	      distance (1 - cosine similarity), (2) build a complete feature graph with edge
	      weights based on pairwise cosine distance, (3) prune edges starting from the
	      least correlated features (processed in ascending order of importance) while
	      maintaining graph connectivity, targeting a graph density of $1.5\times$ the
	      minimum connectivity ratio (capped at 1.0). This approach better captures
	      feature interactions compared to the original method in
	      \citet{zhao2024shapgnewfeatureimportance}.
	\item
	      \textbf{PFI method}:
	      We implement the Permutation Feature Importance (PFI) method
	      \citep{altmann2010permutation} by measuring the drop in a model performance when
	      each feature values are randomly shuffled across samples.
	      The algorithm shuffles each feature several times (5 by default), measures the
	      effect after each shuffle, and averages the results to reduce randomness. Features
	      giving larger drop in the performance when shuffled are deemed more important,
	      providing a model-agnostic evaluation of feature importance without requiring model
	      retraining.
\end{enumerate}

\subsection{Results \& analysis}

\begin{figure}[htb]
	\begin{center}
		\includegraphics[width=0.95\linewidth, clip, trim=0cm 0cm 0cm 0.9cm]{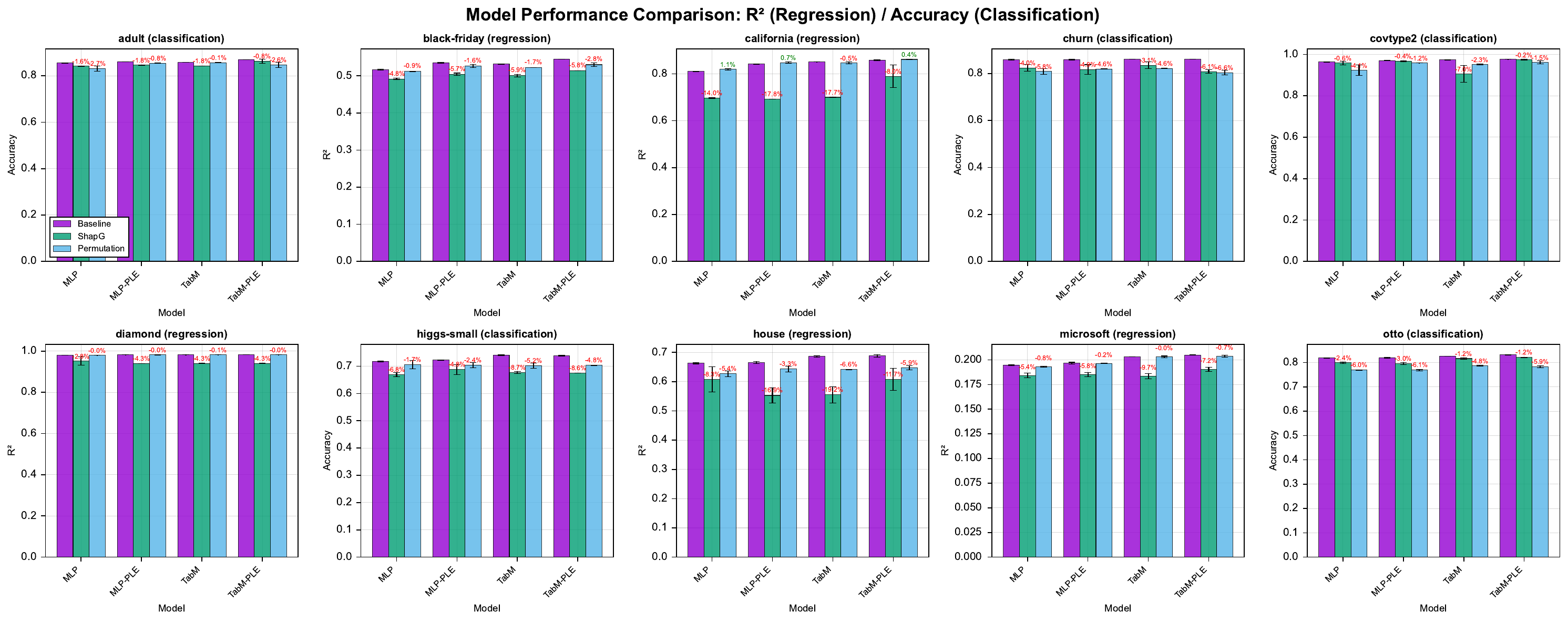}
	\end{center}
	\caption{Model performance ($R^2$ for regression task and accuracy for classification task) across 10 datasets.}
	\label{fig:performance_comparison}
\end{figure}

Under the experimental setup described above, we evaluate performance of the
proposed feature selection pipeline on 10 benchmark datasets. Our analysis
reveals distinct patterns between classification and regression tasks, as shown
in Figure~\ref{fig:performance_comparison}. On the $OX$-axis of each graph, the
four models (MLP, TabM, MLP+PLE, TabM+PLE) are presented. Ten subfigures
correspond to ten datasets presented in Table \ref{tab:dataset_stats}. Accuracy
(for classification tasks) or $R^2$ (for regression tasks) of the model is
presented on the $OY$-axis. The purple columns correspond to the baseline
without feature selection, while the green and light-blue columns correspond to
models using ShapG and Permutation feature selection (top 80\% features),
respectively.

\textbf{Classification tasks.}
For classification tasks (datasets \textit{Adult}, \textit{Churn Modelling},
\textit{Covertype}, \textit{Higgs Small}, \textit{Otto Group Products}), both
feature-selection methods (ShapG and PFI) show consistently strong robustness
to feature reduction. Across all datasets, both ShapG and Permutation selection
of the top 80\% most important features yield accuracy that remains close to
the baseline models trained on all features (purple bars), with typical
degradation of only a few percentage points and an average drop less than 4\%
across all model-dataset combinations. Notably, on datasets such as
\textit{Adult}, both ShapG and PFI maintain performance very close to the
baseline. In some cases (datasets \textit{Covertype} and \textit{Otto Group
	Products}), ShapG significantly outperforms PFI in terms of average accuracy,
whereas on the \textit{Higgs Small} dataset, PFI substantially outperforms
ShapG. For the \textit{Churn Modelling} dataset, both methods exhibit
comparable performance.

\textbf{Regression tasks.}
In regression tasks (datasets \textit{Black Friday}, \textit{California
	Housing}, \textit{House 16H}, \textit{Diamond}, \textit{Microsoft}), feature
reduction affects the two methods very differently: ShapG shows substantial
sensitivity, while the PFI method maintains stable performance. Across all
regression datasets, ShapG exhibits substantial degradation in $R^2$, often
exceeding 10\%, that is observed in datasets \textit{California Housing} and
\textit{House 16H}. In contrast, the PFI method remains much more stable: its
performance drops are consistently small, typically below 6\%, and in some
cases even slightly surpass or remain close to the baseline (e.g., datasets
\textit{California Housing} and \textit{Diamond}). This suggests that for
regression tasks, the PFI method provides a more reliable feature importance
evaluation compared to ShapG.

\begin{figure}[htb]
	\begin{center}
		\includegraphics[width=0.9\linewidth, clip, trim=0cm 0cm 0cm 0.9cm]{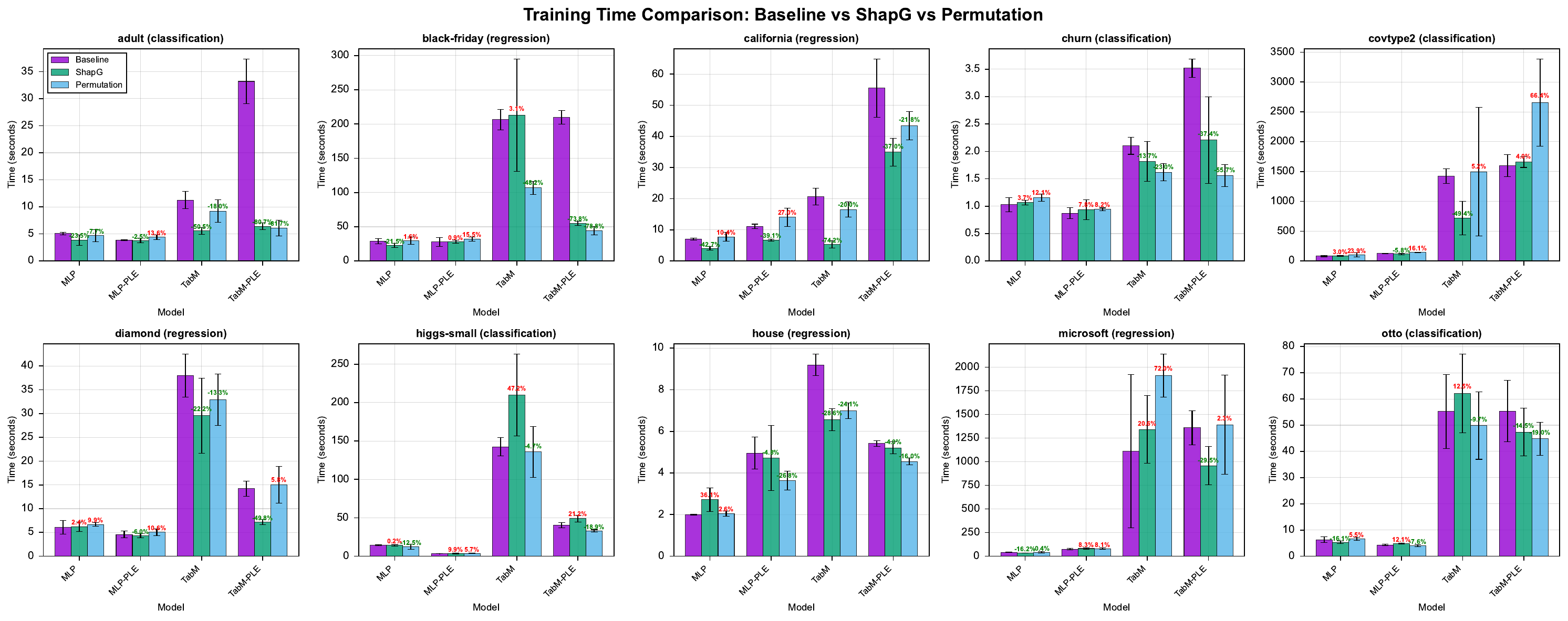}
	\end{center}
	\caption{Training time for different methods across 10 datasets.}
	\label{fig:time_comparison}
\end{figure}

\textbf{Training time.}
As shown in Figure~\ref{fig:time_comparison}, the impact of feature selection
on training time varies across models and datasets. Notably, the 80\% of
feature configurations of both ShapG and PFI do not always lead to smaller
training time compared with the baseline. This arises because the number of
training epochs is not fixed in our setting; instead, we rely on early stopping
with a patience parameter. Consequently, the actual training time is determined
by the convergence dynamics of each model. In some cases, reducing the feature
set accelerates convergence, whereas in others it may slow it down, leading to
longer training time.

\begin{table}[htb]
	\centering
	\caption{Performance across different feature selection methods}
	\label{tab:summary_results}
	\scalebox{0.9}{\begin{tabular}{ll|cccc}
	\hline
	\multirow{2}{*}{XAI Method} & \multirow{2}{*}{Model} & \multicolumn{2}{c|}{Classification} & \multicolumn{2}{c}{Regression}                                      \\
	\cline{3-6}
	                            &                        & Accuracy                            & Time (s)                       & $R^2$            & Time (s)        \\
	\hline
	\multirow{4}{*}{Baseline}   & MLP                    & 0.8426                              & 22.1                           & 0.6329           & 16.9            \\
	                            & MLP-PLE                & 0.8466                              & 27.5                           & 0.6443           & 24.7            \\
	                            & TabM                   & 0.8520                              & 326.6                          & 0.6510           & 277.1           \\
	                            & TabM-PLE               & 0.8554                              & 345.7                          & 0.6561           & 328.8           \\
	\hline
	\multirow{5}{*}{ShapG}      & MLP                    & 0.8182 (-3.1\%)                     & 22.2 (-6.6\%)                  & 0.5865 (-7.1\%)  & 13.9 (-8.4\%)   \\
	                            & MLP-PLE                & 0.8226 (-3.0\%)                     & 26.3 (+4.3\%)                  & 0.5750 (-10.1\%) & 25.0 (-8.1\%)   \\
	                            & TabM                   & 0.8153 (-4.3\%)                     & 199.9 (-10.8\%)                & 0.5760 (-11.3\%) & 318.9 (-20.3\%) \\
	                            & TabM-PLE               & 0.8284 (-3.4\%)                     & 353.0 (-21.5\%)                & 0.6085 (-7.4\%)  & 212.2 (-38.8\%) \\
	                            & Avg.                   & -3.5\%                              & -8.7\%                         & -9.0\%           & -18.9\%         \\
	\hline
	\multirow{5}{*}{PFI}        & MLP                    & 0.8077 (-4.1\%)                     & 25.8 (+4.2\%)                  & 0.6262 (-1.2\%)  & 17.2 (+4.8\%)   \\
	                            & MLP-PLE                & 0.8214 (-3.0\%)                     & 31.7 (+7.2\%)                  & 0.6393 (-0.9\%)  & 27.2 (+6.9\%)   \\
	                            & TabM                   & 0.8237 (-3.4\%)                     & 338.4 (-10.0\%)                & 0.6392 (-1.8\%)  & 415.0 (-6.7\%)  \\
	                            & TabM-PLE               & 0.8197 (-4.3\%)                     & 548.2 (-21.8\%)                & 0.6451 (-1.8\%)  & 299.6 (-21.7\%) \\
	                            & Avg.                   & -3.7\%                              & -5.1\%                         & -1.4\%           & -4.2\%          \\
	\hline
\end{tabular}}
\end{table}

\textbf{Summary.}
Table~\ref{tab:summary_results} highlights complementary strengths of the ShapG
and PFI methods.\footnote{Percentage changes in Table~\ref{tab:summary_results}
	are computed per-dataset and then averaged across all datasets.} ShapG performs
exceptionally well on classification tasks, achieving only an average accuracy
reduction of 3.5\% while delivering the largest reductions in training time
across all models demonstrating its strong efficiency advantage. In contrast,
the PFI method shows remarkable robustness in regression tasks, with an average
drop of only 1.4\% in $R^2$, making it highly reliable when predictive
stability is essential. Taken together, these results show that the ShapG
method offers superior computational benefits, whereas the PFI method excels in
preserving regression performance, providing two complementary and effective
strategies for feature selection.

\section{Conclusions and future work}\label{sec:conclusion}

In this work, we proposed a feature selection pipeline based on game-theoretic
feature importance evaluation. The pipeline employs a $k$-means-based diversity
sampling procedure to obtain a representative subset of observations, thereby
reducing the influence of outliers and enabling more stable feature ranking. We
evaluated two variants of this pipeline, namely, ShapG and PFI approaches,
across ten benchmark datasets using state-of-the-art tabular deep-learning
models. The results show that the proposed pipeline consistently reduces
training cost while maintaining competitive predictive performance. As
summarized in Table~\ref{tab:summary_results}, ShapG offers the largest
reduction in training time with only a modest loss in classification accuracy,
whereas the permutation-based method exhibits higher robustness on regression
tasks and causes only minimal degradation in the corresponding metrics. These
observations indicate that ShapG is a suitable choice when computational
efficiency is prioritized, while the permutation-based method is preferable
when the stability of regression performance is of primary importance.

Our study has several limitations that open opportunities for future research.
First, we evaluated only two feature selection configurations (retaining the
top 80\% of features by importance) using the ShapG and PFI methods.
Comprehensive comparisons with alternative feature selection approaches,
including LIME and other game-theoretic methods with different characteristic
functions, remain to be explored. We can also extend the proposed method by
varying the number of features selected (the percentage of $q$-top features)
for a fixed drop in the model performance. This extension is due to varying the
number of significant features depending of the dataset and deep-learning model
using as a basis.

Additionally, extending this benchmark to industrial datasets, such as those
described in \cite{rubachev2024tabred}, would provide valuable insights on
performance of real-world data analysis.

The proposed diversity sampling method also shows promising results beyond
feature importance evaluation. Preliminary experiments suggest its
effectiveness for training data sampling, in particular, for domains with
heterogeneous data distributions, e.g. wireless network data reported by base
stations (measurement reports). While we have observed encouraging results on
proprietary wireless communication datasets with TabM \citep{gorishniy2024tabm}
and FT-transformer \citep{gorishniy2021revisiting} models, we cannot present
these findings due to confidentiality constraints. Future work should conduct a
systematic comparison between the proposed diversity sampling method and
existing cluster-based approach \citep{axiotis2024data}, and explore these
applications on publicly available datasets.

These findings suggest that feature selection remains a valuable preprocessing
step even for modern deep-learning approaches to tabular data, offering a
practical way to data analysts to balance computational efficiency with respect
to predictive performance.

%

\bibliography{icomp2024_conference}
\bibliographystyle{icomp2024_conference}

\end{document}